\documentclass[final,1p,times,sort&compress]{elsarticle}

\usepackage{amssymb}

\usepackage{graphicx}
\usepackage[T1]{fontenc}
\usepackage{lmodern}
\usepackage{textcomp}
\usepackage{latexsym}
\usepackage{here}

\usepackage{algorithm}
\usepackage{algorithmic}
\usepackage{bm}
\usepackage{mediabb}
\usepackage{amsmath}

\journal{Pattern Recognition}

\begin{document}

\begin{frontmatter}


\title{Denoising random forests}
\author[chubu]{Masaya Hibino}
\author[cslabs]{Akisato Kimura\corref{akisato}}
  \ead{akisato@ieee.org}
\author[chubu]{Takayoshi Yamashita}
\author[chubu]{Yuji Yamauchi}
\author[chubu]{Hironobu Fujiyoshi\corref{hf}}
  \ead{hf@cs.chubu.ac.jp}
\cortext[akisato]{Principal corresponding author}
\cortext[hf]{Corresponding author}
\address[chubu]{%
  Chubu University, Kasugai, Aichi, 487-0027 Japan}
\address[cslabs]{%
  NTT Communication Science Laboratories, Atsugi, Kanagawa, 243-0198 Japan.}

\def\eg{\emph{e.g}\bmvaOneDot}
\def\Eg{\emph{E.g}\bmvaOneDot}
\def\etal{\emph{et al}\bmvaOneDot}
\newcommand{\todo}[1]{{\bf (TODO: {#1})}}

\begin{abstract}
This paper proposes a novel type of random forests called a denoising random forests that are robust against noises contained in test samples.
Such noise-corrupted samples cause serious damage to the estimation performances of random forests, since unexpected child nodes are often selected and the leaf nodes that the input sample reaches are sometimes far from those for a clean sample.
Our main idea for tackling this problem originates from a binary indicator vector that encodes a traversal path of a sample in the forest.
Our proposed method effectively employs this vector by introducing denoising autoencoders into random forests.
A denoising autoencoder can be trained with indicator vectors produced from clean and noisy input samples, and non-leaf nodes where incorrect decisions are made can be identified by comparing the input and output of the trained denoising autoencoder.
Multiple traversal paths with respect to the nodes with incorrect decisions caused by the noises can then be considered for the estimation.
\end{abstract}

\begin{keyword}
  Random forests, regression, denoising autoencoder
\end{keyword}

\end{frontmatter}



\section{Introduction}
\label{sec:intro}

Random forests \cite{Breiman2001}, which constitute an ensemble learning method using multiple decision trees, have a successful history in machine learning, and have many appealing properties that attract a wide range of researchers and engineers.
Their flexible framework makes it possible to apply them not only to classification problems but also to regression, density estimation, semi-supervised learning, and manifold learning \cite{Criminisi2011}.
Random forests have also been experimentally demonstrated to perform well even when handling high-dimensional data problems \cite{Caruana2006}.
In addition, they are inherently distributable on parallel hardware architectures while being close to an ideal learner \cite{Sharp2008,Li2012,DelRio2014,Wakayama2015}.
Random forests are also used for various computer vision tasks, including character recognition \cite{Amit1997}, object recognition \cite{Gall2011}, semantic segmentation \cite{Shotton2008}, and human pose recognition \cite{Shotton2013}.

\begin{figure}[t]
  \begin{center}
    \includegraphics[mediaboxonly,width=0.985\columnwidth]{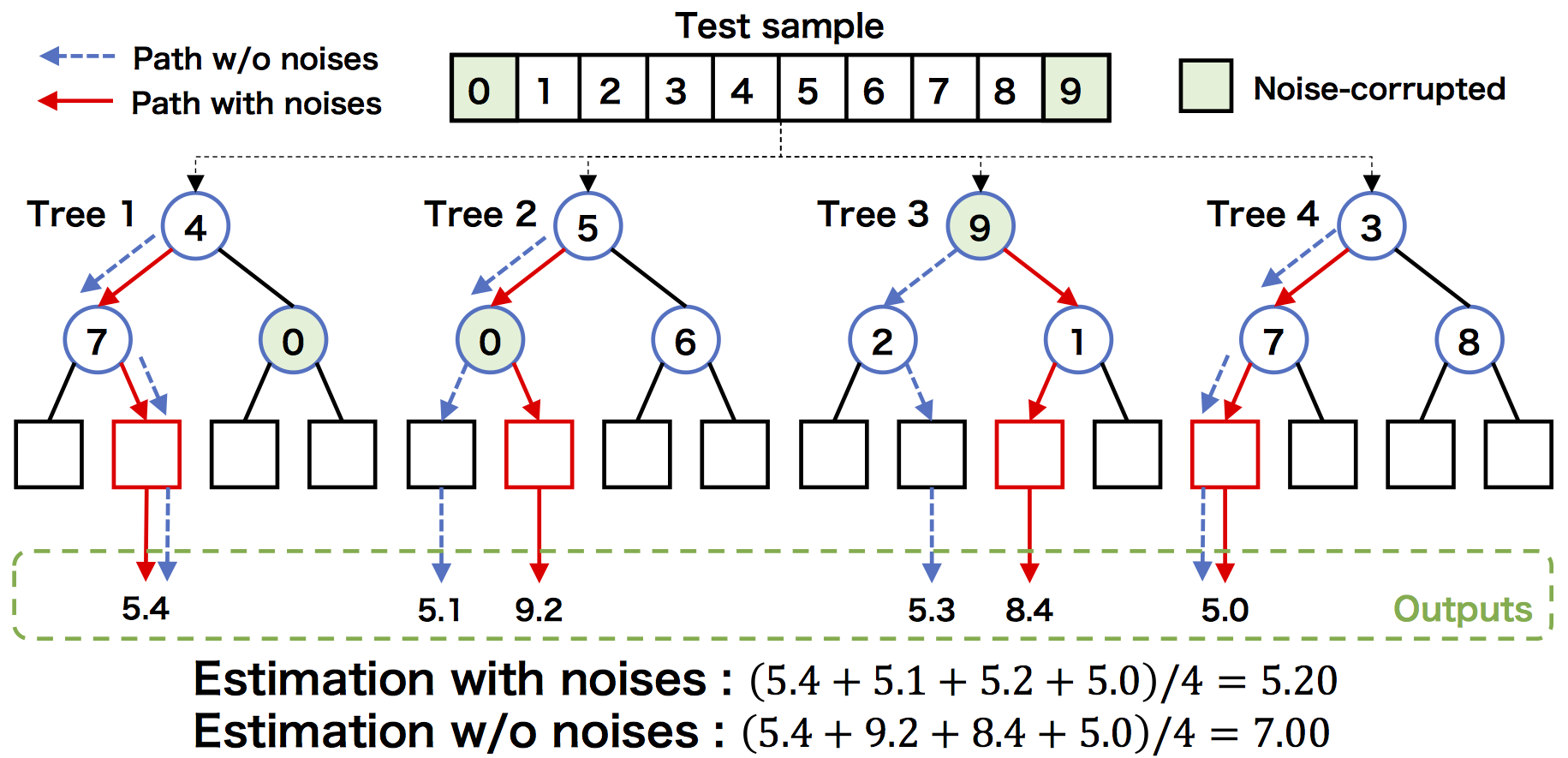}
    \caption{%
      Noise in test samples causes serious damage to the estimation of
      random forests, since leaf nodes are sometimes far from those for
      clean samples.
    }
    \label{fig:problem}
  \end{center}
\end{figure}

One of the fundamental problems with random forests is noise robustness.
Random forests are believed to be robust to noises contained in training samples thanks to their ensemble properties.
However, they do not work well when training samples are clean but test samples are contaminated by noises.
Figure \ref{fig:problem} shows an example where we consider a regression forest with 4 decision trees for 10-dimension features, and several dimensions have been corrupted by noises (filled with green).
Such noise-corrupted samples cause unexpected decisions in nodes (green nodes in Figure \ref{fig:problem}), and the leaf nodes where the sample reached are different or sometimes completely unlike those for clean samples, which results in large estimation errors.
One possible solution is to remove noises from test samples by applying denoising autoencoders \cite{Hinton2006} to test samples.
Zhang and Yan \cite{Zhang2012} discussed two types of prediction errors, bias and variance, in regression forests, and proposed a method for estimating prediction biases that are more difficult to remove than prediction variances.
Also, several previous researches try to detect noise-corrupted test samples as outliers \cite{Fawagreh2015,Ting2008}.
However, those strategies might not work well for very large or bursty noises.

A key insight we are focusing on for managing this issue is that each leaf in a decision tree stores an estimation (e.g. a class histogram for a classification forest and a target value for a regression forest) and a decision tree returns the same estimation result for all the samples felled into the same leaf node.
This readily implies that we can reproduce the same estimation results as those for noiseless samples once we can fix wrong decisions in nodes caused by noises to correct ones.
Recovering noiseless samples from noisy ones remains difficult even if we fully exploit various techniques brought by recent progress on deep learning.
Meanwhile, detecting nodes with incorrect decisions is much easier than those approaches.

\begin{figure}[t]
  \begin{center}
    \includegraphics[mediaboxonly,width=0.985\columnwidth]{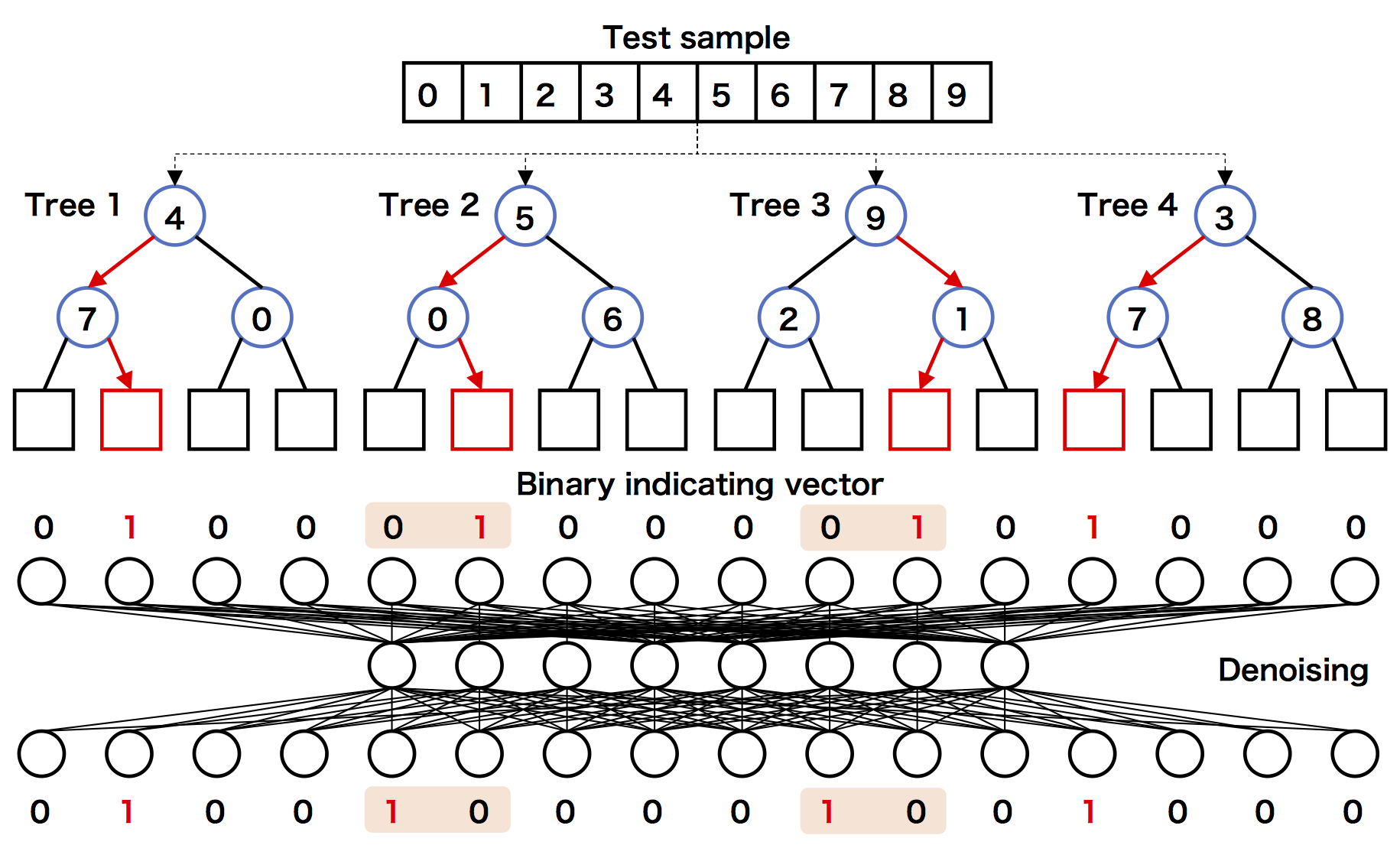}
    \caption{%
      The proposed method eases the negative effects caused by
      noises by introducing autoencoders.
    }
    \label{fig:outline}
  \end{center}
\end{figure}

This paper proposes a novel method for boosting the resistance of random forests to noises in test samples by effectively incorporating the above insight.
We employ binary indicator vectors that have the same dimension as the number of leaf nodes in the forest and that indicate leaf nodes where the input sample has arrived.
Figure \ref{fig:outline} bottom shows an examples of a binary indicator vector.
We can see that a binary indicator vector for an input sample encodes the trajectory of the sample from the root to a leaf node in each decision tree.
We also see that from the nature of decision trees each leaf node in the forest corresponds to a part of the feature space, and thus binary indicator vectors should have a strong correlation.
As shown in Figure \ref{fig:outline}, our proposed method called denoising random forests employs denoising autoencoders to detect and correct noise-corrupted decision nodes.
A denoising autoencoder is trained with indicator vectors produced from clean and noise-contaminated training samples, which enables us to detect nodes where incorrect decisions are made.
If each element of binary indicating vectors can be assumed to be generated from a Bernoulli distribution independently, clean binary indicating vectors from clean training samples cannot be recovered even with denoising autoencoders.
However, as described above, binary indicating vectors have a strong correlation, and thus clean indicating vectors can be recovered with denoising autoencoders.
Once we can detect such nodes, multiple traversal paths, namely the left and right child nodes, with respect to the nodes can then be considered for the estimation.

\section{Regression Forests}
\label{sec:rgForest}

Our proposed method can be applied to any type of decision forests, however, we here concentrate on the use of regression forests \cite{Criminisi2011}, since they are more sensitive to noises than any other types of forests.
In the following, we first describe a procedure of regression forests.

\def\x{\bm{x}}
\def\y{\bm{y}}
\def\z{\bm{z}}
\def\X{\bm{X}}
\def\Y{\bm{Y}}
\def\Z{\bm{Z}}
\def\bB{\mathbb B}
\def\bR{\mathbb R}
\def\cN{\mathcal N}
\def\cS{\mathcal S}
\def\cT{\mathcal T}

It begins by creating $T$ subsets $\X_1, \X_2, \ldots, \X_T$ of training samples $\X=\{(\x_1,y_1),(\x_2,y_2),\ldots,(\x_M,y_M)\}$, where $\x_m=(x_{m1},\ldots,x_{mD})^{\top}\in\bR^D$ and $y_m\in\bR$ are a feature vector and the target value of the $m$-th sample, respectively.
Random forests usually adopt bootstrap sampling to create subsets, and thus subsets $\X_1, \X_2, \ldots, \X_T$ might share several samples.
Next, the $t$-th decision tree $\cT_t$ for regression is built from the $t$-th subset $\X_t$ as training samples.
Each node $\cN_{tn}$ of a tree $\cT_t$ selects a feature dimension $d_{tn}$ and a threshold $\theta_{tn}$ that maximizes the following objective function $s(\cS_{tn},d_{tn},\theta_{tn})$, depending on the index set $\cS_{tn}$ of samples that arrive at the $n$-th node $\cN_{tn}$, the selected dimension $d_{tn}$ and the threshold $\theta_{tn}$:
\begin{align*}
  & s(\cS_{tn},d_{tn},\theta_{tn})
     = \sum_{m\in \cS_{tn}} (y_m-\bar{y}(\cS_{tn}))^2
    - \sum_{i=L,R}\left[\sum_{m\in \cS_{tn}}
         \left\{y_m-\bar{y}(\cS_{tn}^{(i)}(d_{tn},\theta_{tn}))\right\}^2\right],\\
  & \cS_{tn}^{(L)}(d,\theta)
     = \{m\in\cS_{tn} | x_{md}\le\theta\},\\
  & \cS_{tn}^{(R)}(d,\theta)
     = \{m\in\cS_{tn} | x_{md}>\theta\},
\end{align*}
where $\bar{y}(\cS)$ is the average target value $y$ across all the samples in the set $\cS$.
Then, all the samples in $\cS_{tn}$ are divided into the left and right child nodes $\cN_{tn}^{(L)}$ and $\cN_{tn}^{(R)}$ whose sample sets are $\cS_{tn}^{(L)}(d_{tn},\theta_{tn})$ and $\cS_{tn}^{(R)}(d_{tn},\theta_{tn})$, respectively.
If the node $\cN_{tn}$ reaches a pre-defined maximum depth, a target estimate $\hat{y}_{tn}$ for samples that arrive at the node $\cN_{tn}$ is computed by
\begin{align*}
  & \hat{y}_{tn}
    = \left(\sum_{m\in\cS_{tn}}y_m\right) / \left(\sum_{m\in\cS_{tn}}1\right)
\end{align*}

In the testing stage, each decision tree $\cT_t$ receives a test sample $\x_{\mbox{test}}$, transfers it to the leaf node according to the splitting functions $s(\cdot)$ of nodes, and returns a target estimate $\hat{y}_t$ stored in the leaf node.
The final output $\hat{y}$ is the average over the outputs $\hat{y}_1(\x_{\mbox{test}}), \ldots, \hat{y}_T(\x_{\mbox{test}})$ obtained from a test sample $\x_{\mbox{test}}$ at the trees in the forest, as follows:
\begin{align*}
  \hat{y} &= \sum_{t=1}^T\hat{y}_t(\x_{\mbox{test}})/T
\end{align*}

As described in the introduction and Figure \ref{fig:problem}, noises contained in test samples cause undesirable decisions in nodes, which results in large regression errors.
We empirically evaluated regression errors with respect to the amount of noises in test samples, using a public dataset named Concrete Compressive Strength dataset \cite{Yeh1998} in the UCI Machine Learning Repository.
In this experiment, we used 25 trees with depth of 4 and each node of trees in the forest checked a single dimension for splitting samples.
Table \ref{tab:hh} gives the result.
This indicates that noises brought a significant impact on regression errors and this negative impact rapidly increased as the noises increase.

\begin{table}[t]
  \centering
  \caption{%
    Noises in test samples brought a significant impact on regression errors and this negative impact increased rapidly as the noises increase (see Section \ref{sec:rgForest} for the detail).
  }
  \label{tab:hh}
  {\footnotesize
  \begin{tabular}{c|c|c|c|c|c|c|c} \hline
    S/N [\%]   &  0   & 12.5 & 25   & 37.5 & 50   & 62.5 & 75   \\ \hline
    Error [\%] & 15.2 & 16.3 & 22.4 & 23.7 & 26.9 & 29.1 & 30.9 \\ \hline
  \end{tabular}
  }
\end{table}

\section{Proposed method}
\label{sec:proposed}

We propose a novel method called denoising random forests that are robust to noises contained in test samples.

\subsection{Main idea}
\label{sec:proposed:idea}

Before describing the framework of the proposed method, we introduce the main idea behind the proposed method and present a preliminary experimental result that supports our idea.

Our main idea originates from binary indicator vectors $\z\in\bB^N$ that have the same dimension as the number of leaf nodes in the forest, $N=\sum_{t=1}^T N_t$, and that indicate leaf nodes where the input sample arrived (see Figure \ref{fig:outline}).
We here note that each leaf node in a decision tree directly corresponds to a region in the feature space, and thus a binary indicator vector $\z_t\in\bB^{N_t}$ obtained from a single decision tree $\cT_t$ roughly encodes the position of a given feature $\x$ in the feature space.
This readily implies that the representation of the feature space by binary indicator vectors obtained from whole the forest is becoming redundant and the binary indicator vector has a strong correlation, if there are a sufficiently large number of trees in the forest and each tree is sufficiently deep.

\begin{figure}[t]
  \begin{center}
    \includegraphics[mediaboxonly,width=0.785\columnwidth]{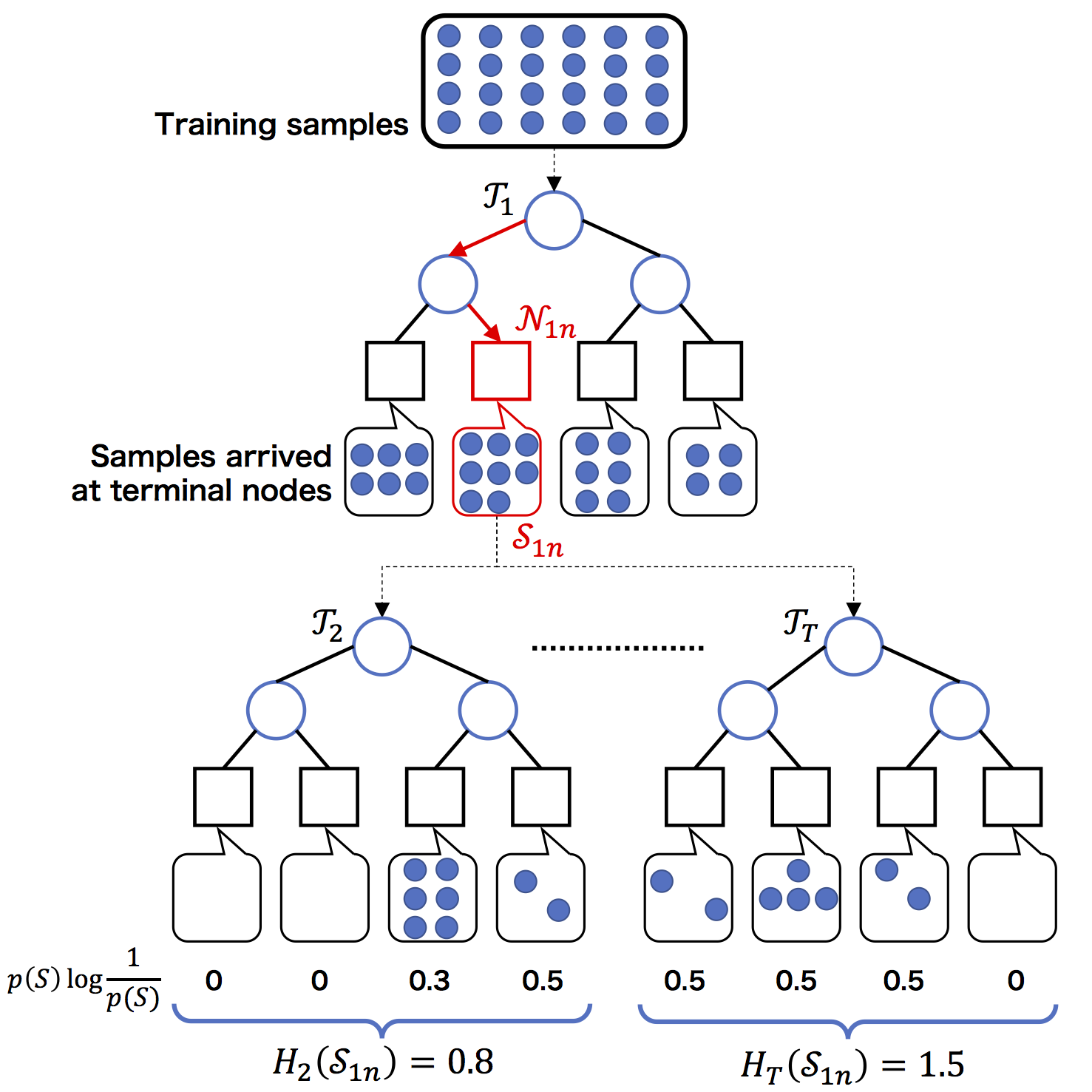}
    \caption{%
      Computing cross entropy of trees in the forest
      (See Section \ref{sec:proposed:idea} for the detail)
    }
    \label{fig:entropy}
  \end{center}
\end{figure}

\begin{figure}[t]
  \begin{center}
    \includegraphics[mediaboxonly,width=0.985\columnwidth]{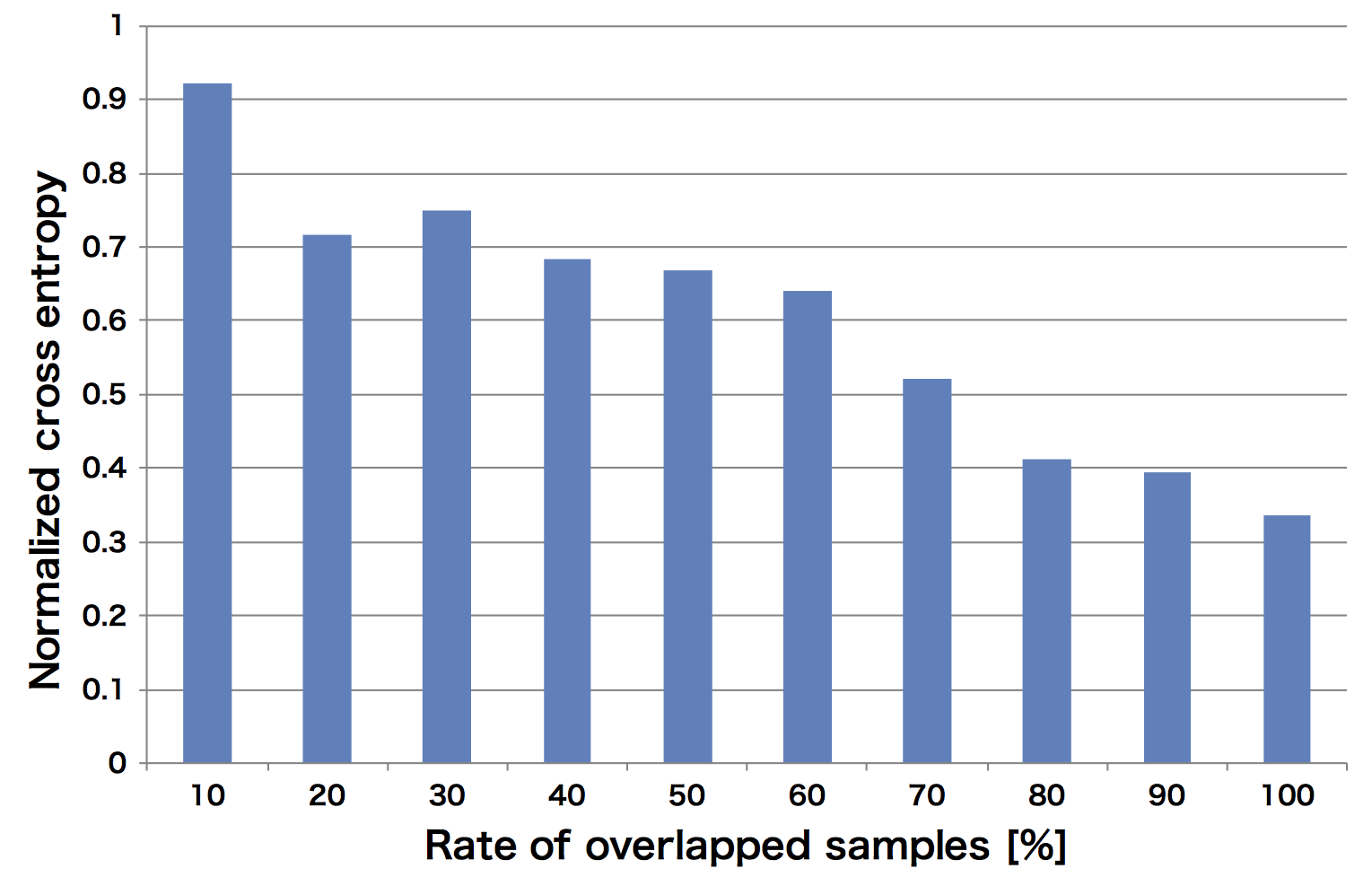}
    \caption{%
      Cross entropy for varying ratios of overlapping samples
    }
    \label{fig:entropy_graph}
  \end{center}
\end{figure}

To confirm the above assumption, we performed a preliminary experiment to compute cross correlations among decision trees.
As a measure for evaluating the correlations, we used cross entropy that can be computed from an aggregation of binary indicator vectors.
More specifically, we follow the scheme below (see also Figure \ref{fig:entropy}):
First, the entropy $H_t(\cS)$ of the $t$-th tree $\cT_t$ for a given sample set $\cS$ is calculated as follows:
\begin{align*}
  & H_t(\cS)
    = -\sum_{n=1}^{N_t} P(\cS_{tn}(\cS);\cS)\log P(\cS_{tn}(\cS);\cS),\\
  & P(\cS_{tn}(\cS);\cS)
    = |\cS_{tn}(\cS)| / |\cS|,\nonumber
 \end{align*}
where $N_t$ is the number of leaf nodes in the $t$-th tree $\cT_t$ and $\cS_{tn}(\cS)$ is a subset of $\cS$ whose samples reach the $n$-th leaf node $\cN_n$ in the $t$-th tree $\cT_t$.
We can see that a small $H_t(\cS)$ means that the outputs of the $t$-th tree $\cT_t$ are highly biased for samples in the set $\cS$.
The cross entropy $C_{t,t'}(\cS)$ between the $t$-th and $t'$-th trees $\cT_t$ and $\cT_{t'}$ for a given sample set $\cS$ and its average $C_t(\cS)$ for the $t$-th tree $\cT_t$ over all the other trees can be computed as follows:
\begin{align*}
  C_{t,t'}(\cS) &= \frac{1}{N_t}\sum_{n=1}^{N_t} H_{t'}(\cS_{tn}(\cS)),\\
  C_t(\cS) &= \frac{1}{T-1}\sum_{t'\neq t}C_{t,t'}(\cS) \nonumber
\end{align*}
We can see that a small $C_{t,t'}(\cS)$ means that samples $\cS_{tn}(\cS)$ felled into a specific leaf node $\cN_{tn}$ in the tree $\cT_t$ fall into only a few nodes in the tree $\cT_{t'}$ and thus the outputs of two trees $\cT_t$ and $\cT_{t'}$ are highly correlated with each other.

\begin{figure}[ht]
  \begin{center}
    \includegraphics[mediaboxonly,width=0.985\columnwidth]{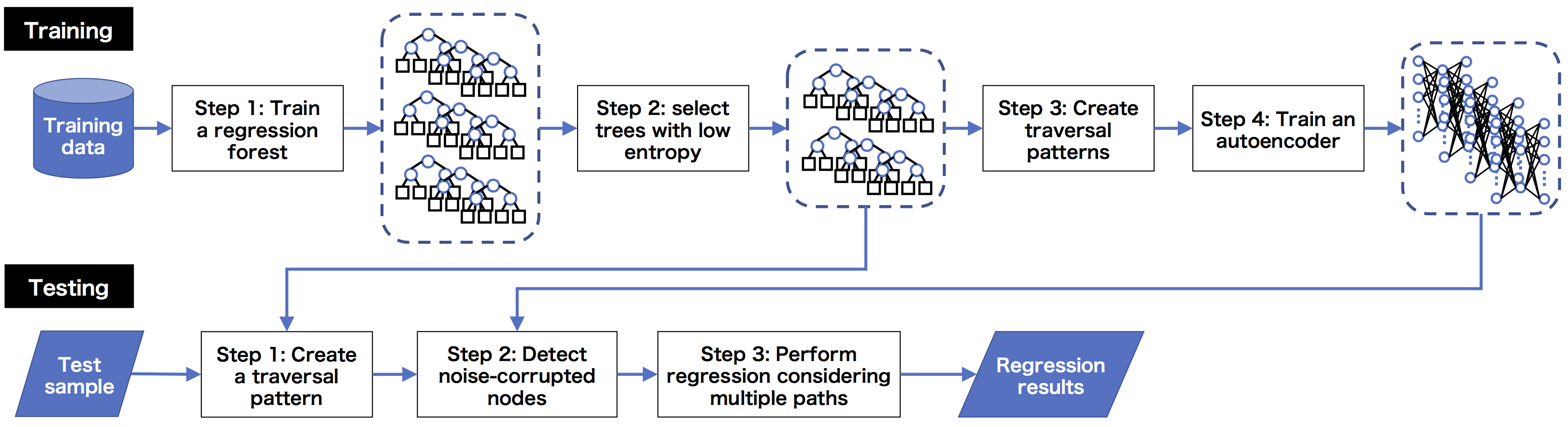}
    \caption{Flow of proposed method}
    \label{fig:flow}
  \end{center}
\end{figure}

The cross entropy $C_{t,t'}(\cS)$ is influenced by the ratio of overlapping samples used for training the $t$-th and $t'$-th trees, since random forests usually adopt bootstrap sampling to generate subsets of training samples, as shown in Section \ref{sec:rgForest}.
Figure \ref{fig:entropy_graph} shows normalized mean cross entropy $C_t(\cS)$ averaged over all the trees in the forest when controlling the ratio of overlapping training samples for bootstrap sampling.
Almost all the experimental settings were the same as Section \ref{sec:rgForest}, and we prepared another subset of samples for computing cross entropy, which are different from the training samples.
This result indicates that the increase in the ratio contributed greatly to the decrease in the cross entropy among trees.
We do not usually control the ratio of overlapping training samples for bootstrap sampling when training random forests, namely, all the training samples are used for bootstrap sampling in general.
Thus, the experimental result shown in Figure \ref{fig:entropy_graph} supports our assumption that trees in the forest are correlated with each other.

The other important characteristic of binary indicator vectors is that each vector represents a traversal path of decision trees.
In other words, a target estimate $\hat{y}$ is uniquely determined by a binary indicator vector $\z$.
\begin{align}
  \hat{y} &= f(\z) = f(g(\x)).
\end{align}
This means that if indicator vectors obtained from noise-free input features can somehow be recovered, appropriate traversal paths can also be recovered and then the estimation performance can be improved without the need for any explicit noise reduction techniques for input samples.

\subsection{Framework}
\label{sec:proposed:framework}

Figure \ref{fig:flow} shows an overview of our proposed method called denoising random forests.
We again note that we assume that all the training samples are clean but test samples might be corrupted by noises.

First in the training stage, a regression forest is trained from a given set of training samples.
See Section \ref{sec:rgForest} for the procedure of training regression forests.
Next, the mean cross entropy $C_t(\cS)$ defined in Section \ref{sec:proposed:idea} is computed for each tree, and trees with low entropy scores are selected to form a refined regression forest.
In general, random forests are expected to have high-entropy trees for improving generalization performances.
On the other hand, this step for selecting low-entropy trees is significant to improve the noise robustness.
We will show the effectiveness of this step experimentally in Section \ref{sec:exp:select}.
Binary indicator vectors each of which represents a traversal path of a training sample are then computed from clean and noisy training samples to train denoising autoencoders.
As with usual denoising autoencoders, we can generate noisy training samples from clean training samples.

In the testing stage, a binary indicator vector is first extracted from a given test sample with the trained regression forest, to be fed into the trained denoising autoencoder.
We can detect nodes that produce incorrect decisions caused by noises from the denoising autoencoder.
The details how to detect noise-corrupted nodes will be described in Section \ref{sec:proposed:denoising}.
We can estimate a possibly correct traversal path from the detected nodes.
Once we can estimate a corrected traversal path, both the original and recovered traversal paths can then be considered for the regression.
The details will be described in Section \ref{sec:proposed:multipath}.

\subsection{Recovering traversal paths}
\label{sec:proposed:denoising}

\begin{figure}[t]
  \begin{center}
    \includegraphics[mediaboxonly,width=0.985\columnwidth]{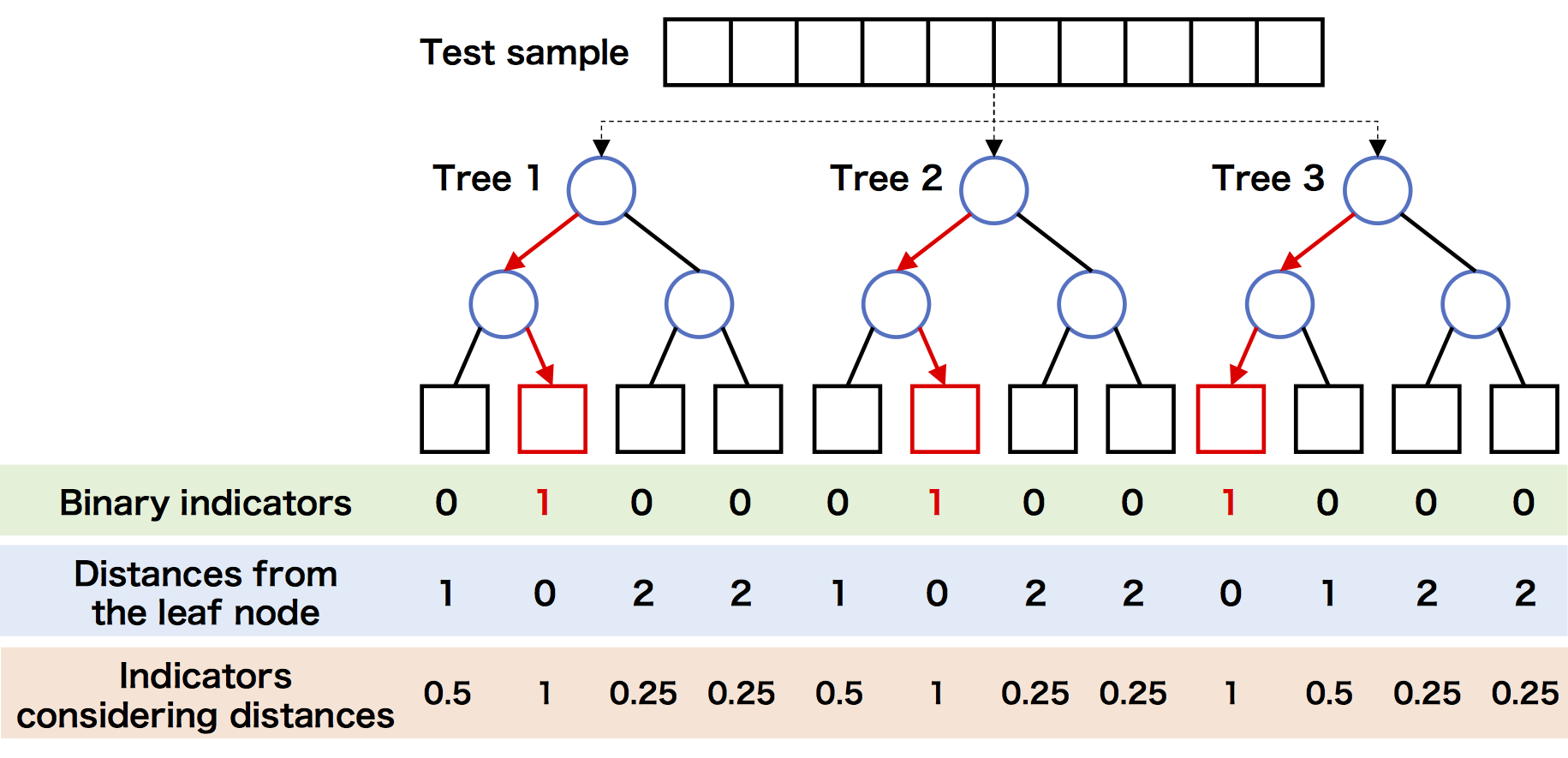}
    \caption{%
      Computing indicator vectors considering inter-node distances.
      See Section \ref{sec:proposed:denoising} for the detail.
    }
    \label{fig:indicator}
  \end{center}
\end{figure}

\begin{figure}[t]
  \begin{center}
    \includegraphics[mediaboxonly,width=0.985\columnwidth]{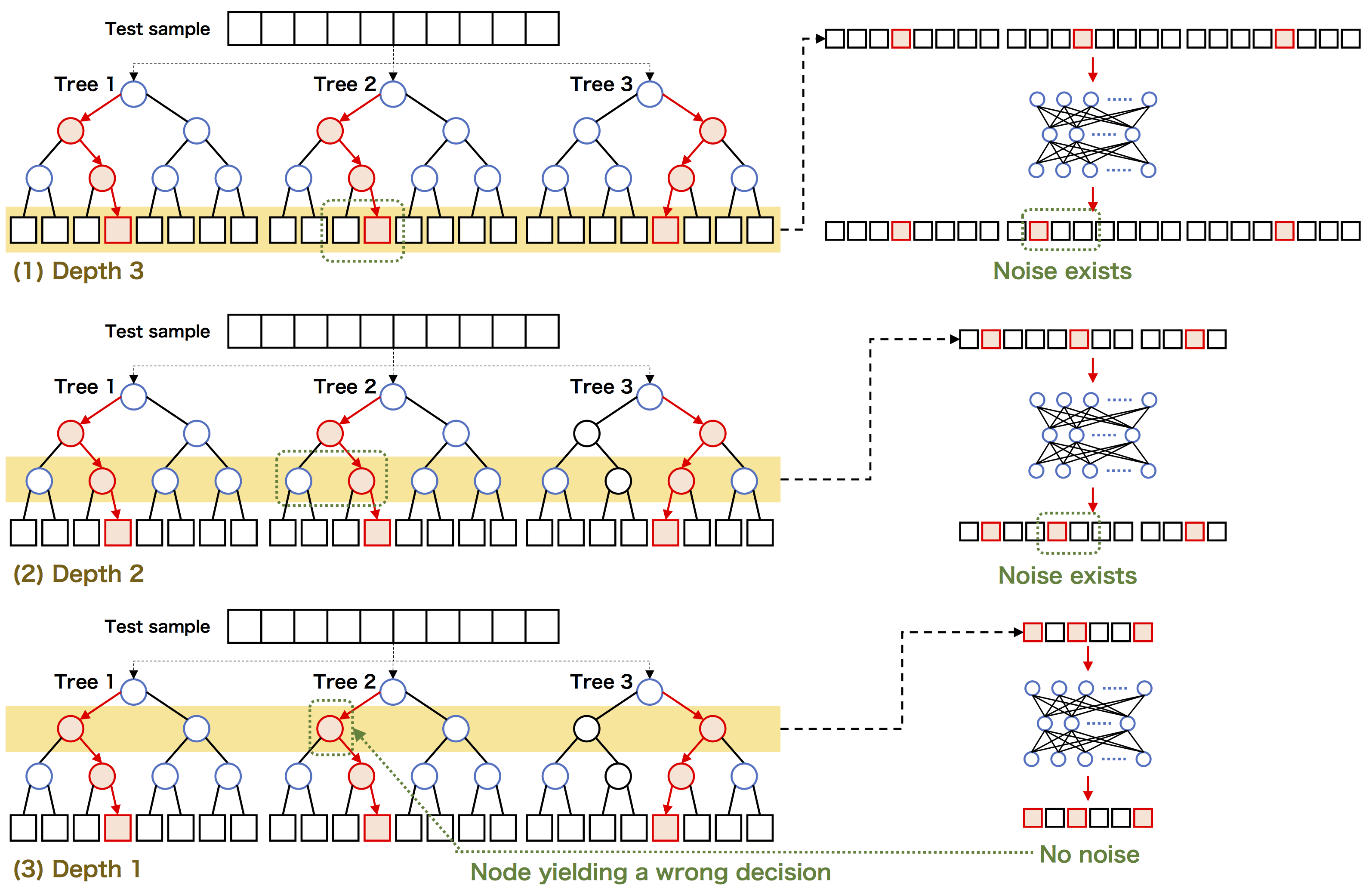}
    \caption{%
      Recovering traversal paths from leaves to root.
      See Section \ref{sec:proposed:denoising} for the detail.
    }
    \label{fig:multiLayer}
  \end{center}
\end{figure}

A denoising autoencoder is trained with indicator vectors produced from clean and noisy samples, where we assume that all the training samples are noiseless.
In this paper, we used a simple architecture for a denoising autoencoder that consists of an input layer, a bottleneck layer and an output layer, and the number of units for the bottleneck layer is half of the number of units in the input and output layers (i.e. the number of leaf nodes in the forest).
A binary indicator vector $\z=g(\x)$ is highly sparse especially with deep trees, which makes it difficult to train a denoising autoencoder.
Therefore, we develop another type of indicator vector $\tilde{\z}=\tilde{g}(\x)$ that encodes inter-node distances.
As shown in Figure \ref{fig:indicator}, each element $\tilde{z}_n$ of the new indicator vector $\tilde{\z}$ that corresponds to the leaf node $\cN_n$ is 1 if the node $\cN(\x)$ where the test sample $\x$ arrives at the node $\cN_n$, and otherwise an element $\tilde{z}_n$ can be computed by the inverse of distance from the leaf node $N(\x)$ where the test sample $\x$ arrives.
\begin{align}
  \tilde{z}_n &= \left\{
    \begin{array}{ll}
      1 & \mbox{if } \cN_n=\cN(\x)\\
      (d(\cN_n, \cN(\x))/2)^{-1} & \mbox{otherwise}
    \end{array}
  \right.
\end{align}
We can improve the performance of recovering traversal paths by preparing denoising autoencoders not only for the leaf depth but also for intermediate node depths.
By doing this, we can identify any nodes in any layers with possibly incorrect decisions.

In the testing stage, traversal paths are recovered with the trained denoising autoencoder from indicator vectors that might be corrupted by noises, by following the steps below (see also Figure \ref{fig:multiLayer}):
A proposed indicator vector $\tilde{\z}=\tilde{g}(\x)$ is first computed from a test sample $\x$, and is passed through the trained autoencoder to obtain a refined indicator vector $\hat{\z}$.
The refined indicator vector is then binarized so that only one element per tree has a positive value, and a refined traversal path can then be recovered from this binarized indicator vector.
Nodes with incorrect decisions can be identified by comparing the differences between the original and refined traversal paths.
When introducing multiple autoencoders depth-wise, the recovery process starts from the leaf depth and moves up to shallower depths, as shown in Figure \ref{fig:multiLayer}.

\subsection{Estimation considering multiple paths}
\label{sec:proposed:multipath}

\begin{figure}[t]
  \begin{center}
    \includegraphics[mediaboxonly,width=0.985\columnwidth]{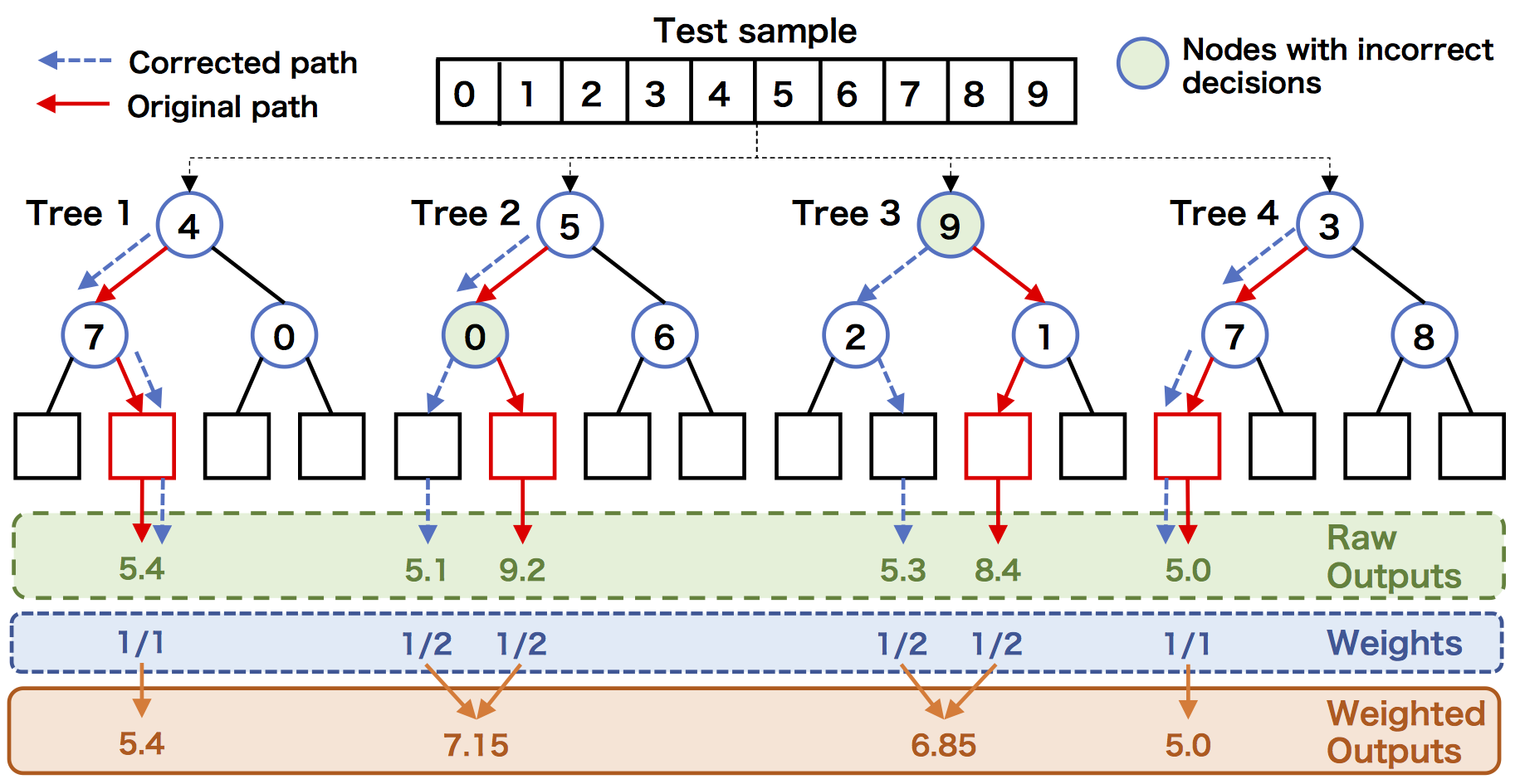}
    \caption{%
      Regression forest considering multiple paths
    }
    \label{fig:multipath}
  \end{center}
\end{figure}

If we assume that we can perfectly recover the true traversal paths $\z$ with denoising autoencoders, we can simply replace the original (possibly noise-corrupted) traversal paths with a refined one.
However, denoising autoencoders may sometimes fail and undesirable traversal paths may recover.
Instead, we propose an estimation process that considers multiple, namely the original and refined, traversal paths.
Figure \ref{fig:multipath} shows an outline of the proposed multiple path estimation.
We have already obtained nodes (filled with green in Figure \ref{fig:multipath}) that might yield incorrect decisions by employing denoising autoencoders (see Section \ref{sec:proposed:denoising}).
Multiple traversal paths are generated from the obtained nodes with incorrect decisions, and an estimation result can be obtained from each traversal path.
\def\cP{\mathcal P}
The final estimation result $\hat{y}$ is a weighted sum of estimation results each of which is obtained from an individual traversal path.
\begin{align*}
  \hat{y} &= \sum_{t=1}^T\sum_{p\in\cP}w_p\hat{y}_{tn_p}/T,\\
  w_p &= 2^{-k_p}
\end{align*}
where $\cP$ is a set of all the traversal paths generated by the above procedure, $n_p$ is an index of a leaf node of the traversal path $p$, $w_p$ is a weight of the traversal path $p$ and $k_p$ is the number of nodes with incorrect decisions in the traversal path $p$.

\section{Evaluation Experiments}
\label{sec:exp}

We performed several experiments to evaluate the effectiveness of our proposed method.

We used five public datasets in the UCI Machine Learning Repository, namely Physicochemical Properties of Protein Tertiary Structure (abbr. Protein), Concrete Compressive Strength (Concrete) \cite{Yeh1998}, Geographical Origin of Music (Music) \cite{Zhou2014}, Energy Efficiency (Energy) \cite{Tsanas2012}, and Airfoil Self-Noise (Airfoil) \cite{Brooks1989}.
We created noisy test samples by intentionally injecting noises into the original features: meanwhile we used the original data for model (namely, regression forests and denoising autoencoders) training.
Although several types of noises can be considered, we here replaced several elements of the samples with the average value across all the samples.
All the parameters of the decision forests were optimized to obtain the best regression performance for the baseline.
More specifically, the number and depth of trees for each dataset were set as follows:
Protein: 40 trees with depth 4, Concrete: 15 trees with depth 4, Music: 35 trees with depth 4, Energy: 20 trees with depth 4, Airfoil: 20 trees with depth 4.

\subsection{Denoising performances}
\label{sec:exp:denoising}

\begin{figure}[t]
  \begin{center}
    \includegraphics[mediaboxonly,width=0.985\columnwidth]{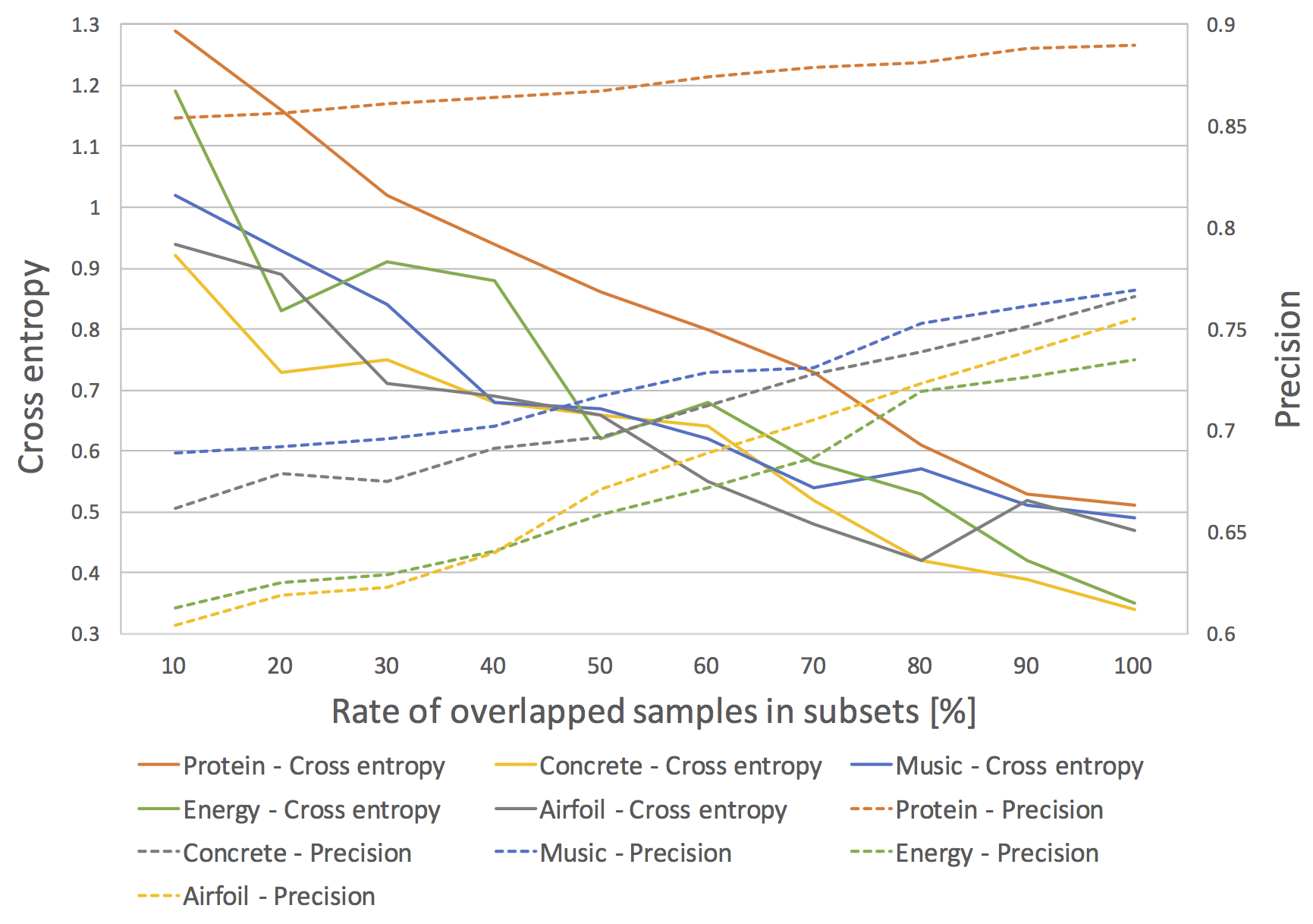}
    \caption{%
      Denoising performance,
      see Section \ref{sec:exp:denoising} for the detail.
    }
    \label{fig:exp_denoise}
  \end{center}
\end{figure}

First, we evaluated the performance of our noise correction with denoising autoencoders described in Section \ref{sec:proposed:denoising}.
As measures for evaluation, we used (1) the precision rate, namely the ratio of samples from which the correct indicator vectors could be recovered, and (2) the cross entropy between the ground-truth and recovered indicator vectors.
We have to note that we adopt a standard definition of cross entropy for this evaluation, and it is different from that shown in Section \ref{sec:proposed:idea}.
The precision rate is rather conservative since it gains only if an estimated indicator vector is strictly equal to the ground-truth.
Meanwhile, the cross entropy enables us to measure a detailed difference between an estimated and ground-truth indicator vector.

Fig. \ref{fig:exp_denoise} shows the results, which indicate that the denoising performance of the proposed method improved (lower cross entropy scores: dashed lines and higher precision rates: solid lines) as the number of overlapping samples in the training subsets increased.
We can also see that the proposed method recovered an average of 75-90\% of the true indicator vectors.

\subsection{Regression performances}
\label{sec:exp:regression}

\begin{figure}[t]
  \begin{center}
    \includegraphics[mediaboxonly,width=0.485\columnwidth]{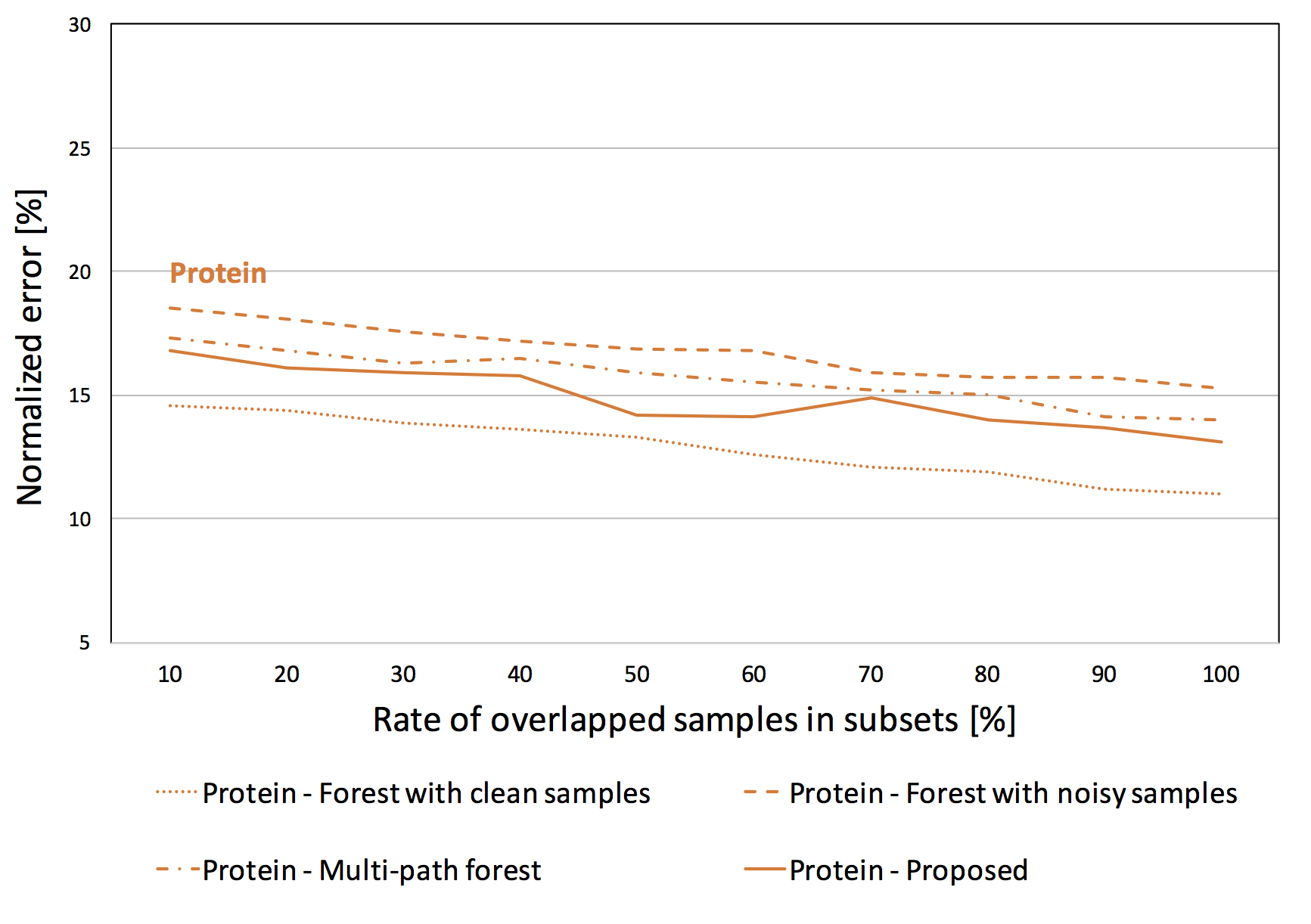}
    \includegraphics[mediaboxonly,width=0.485\columnwidth]{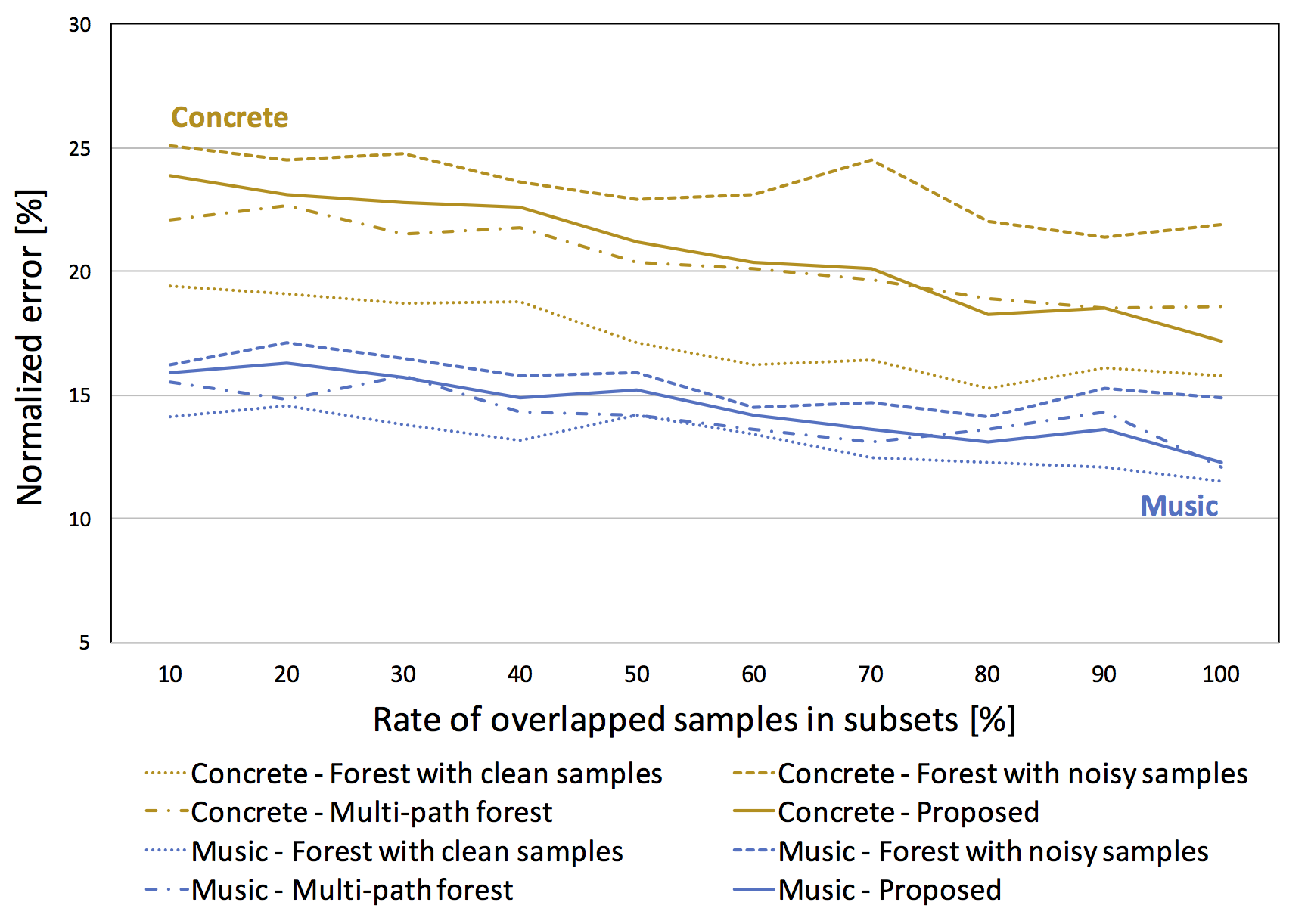}
    \includegraphics[mediaboxonly,width=0.485\columnwidth]{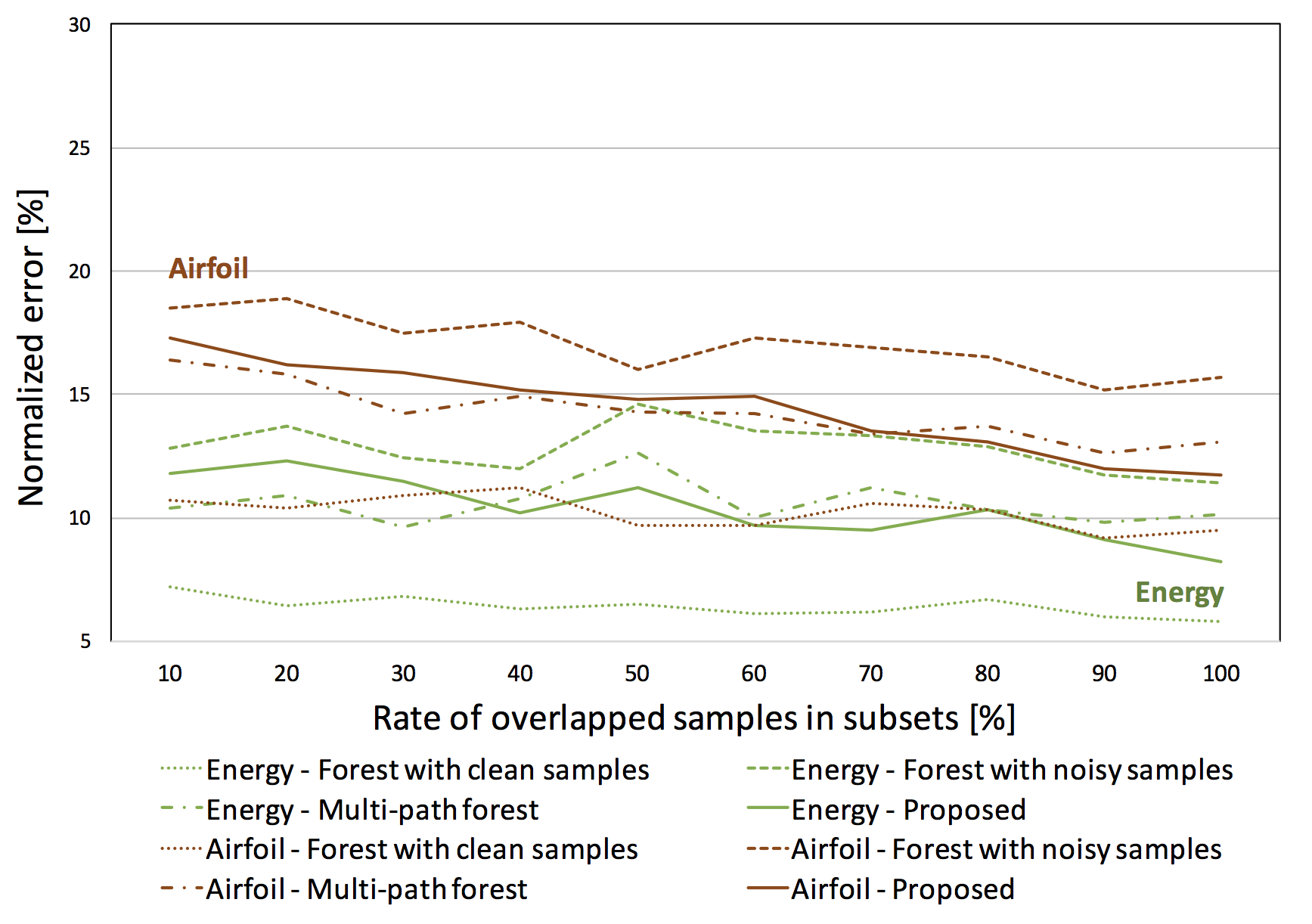}
    \caption{%
      Regression performances for 
      (top left) Protein dataset, (top right) Concrete (yellow) and Music (blue) datasets,
      (bottom) Energy (green) and Airfoil (brown) datasets,
      see Section \ref{sec:exp:regression} for the detail.
    }
    \label{fig:exp_regress}
  \end{center}
\end{figure}

Next, we evaluated the regression performances of the proposed method.
As a measure for evaluations, we used the $L_1$-distance normalized by the ground-truth.
We compared the proposed method with (1) a standard regression forest with clean test samples as an oracle, (2) a standard regression forest with noisy test samples as a baseline, and (3) a multi-path regression forest with noisy test samples and known ground-truth noise patterns.
Figure \ref{fig:exp_regress} show the results.
The naive regression forest provided the best regression performances if the test samples were noise free (dotted lines). However, the performance worsened for the noisy test samples (dashed lines).
The multi-path regression forest (dotted and dashed lines) recovered the regression performance if we knew the true noise patterns, however in real-world applications we cannot obtain such ground-truth noise patterns.
On the other hand, the proposed method (solid lines) exhibited regression performance that was better than the naive regression forest and comparable to that of the multi-path regression forest with true noise patterns, even though we did not know the true noise patterns.

\begin{figure}[t]
  \begin{center}
    \includegraphics[mediaboxonly,width=0.985\columnwidth]{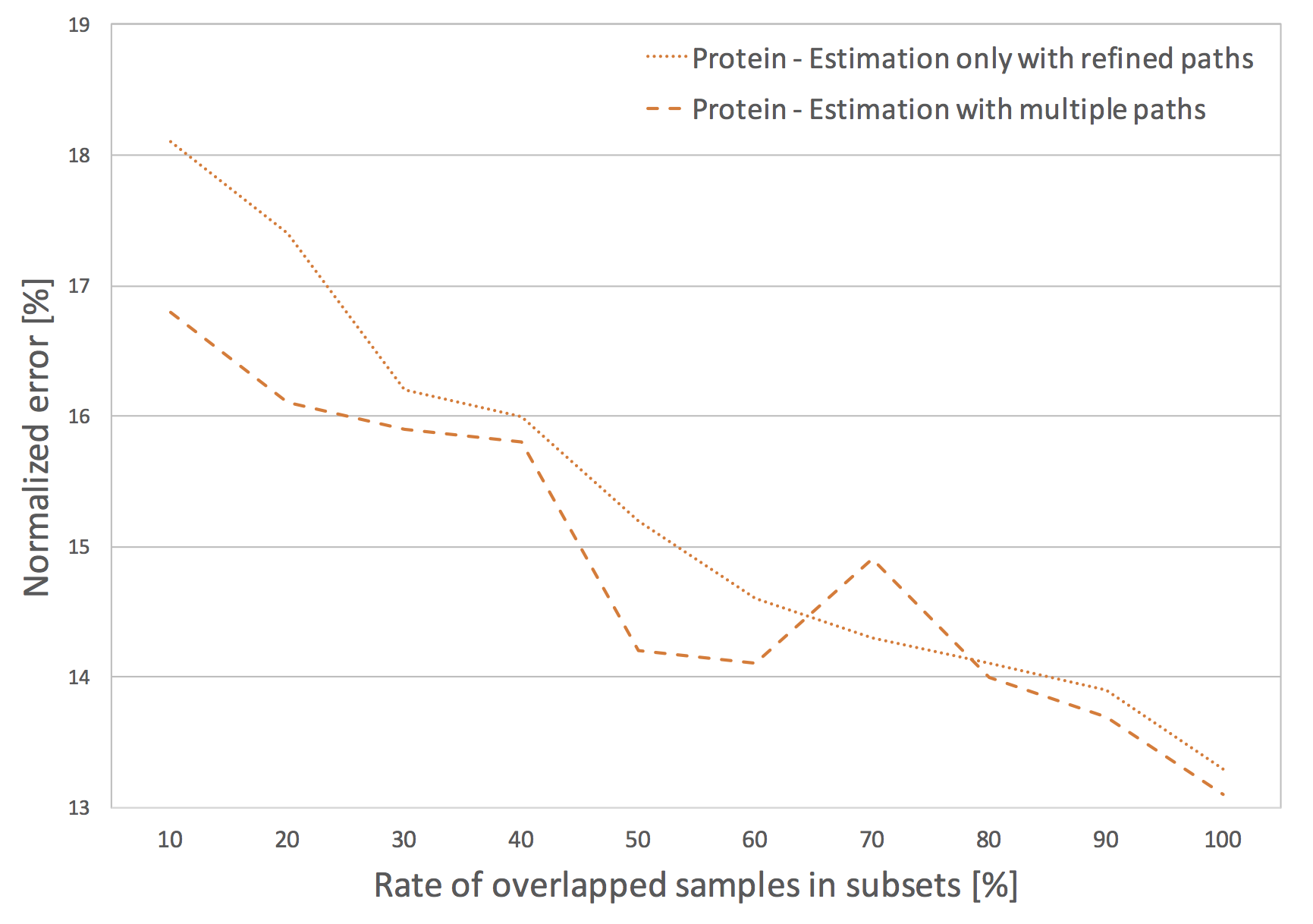}
    \caption{%
      Effectiveness of the proposed multiple path estimation
    }
    \label{fig:exp_multipath}
  \end{center}
\end{figure}

As shown in Figure \ref{fig:exp_denoise}, denoising autoencoders cannot perfectly recover the true traversal paths.
This implies that we should not completely rely on refined traversal paths and we should adopt multiple path estimation proposed in Section \ref{sec:proposed:multipath}.
To confirm the above discussion, we perform another experiment for comparing (1) the regression forest with only refined traversal paths and (2) the proposed method utilizing multiple paths.
Figure \ref{fig:exp_multipath} shows the result for Protein dataset, where solid and dashed lines correspond to the regression forest with only refined traversal paths and the proposed method, respectively.
As you can see in this figure, this experimental result supports our suggestion.

\begin{figure}[t]
  \begin{center}
    \includegraphics[mediaboxonly,width=0.485\columnwidth]{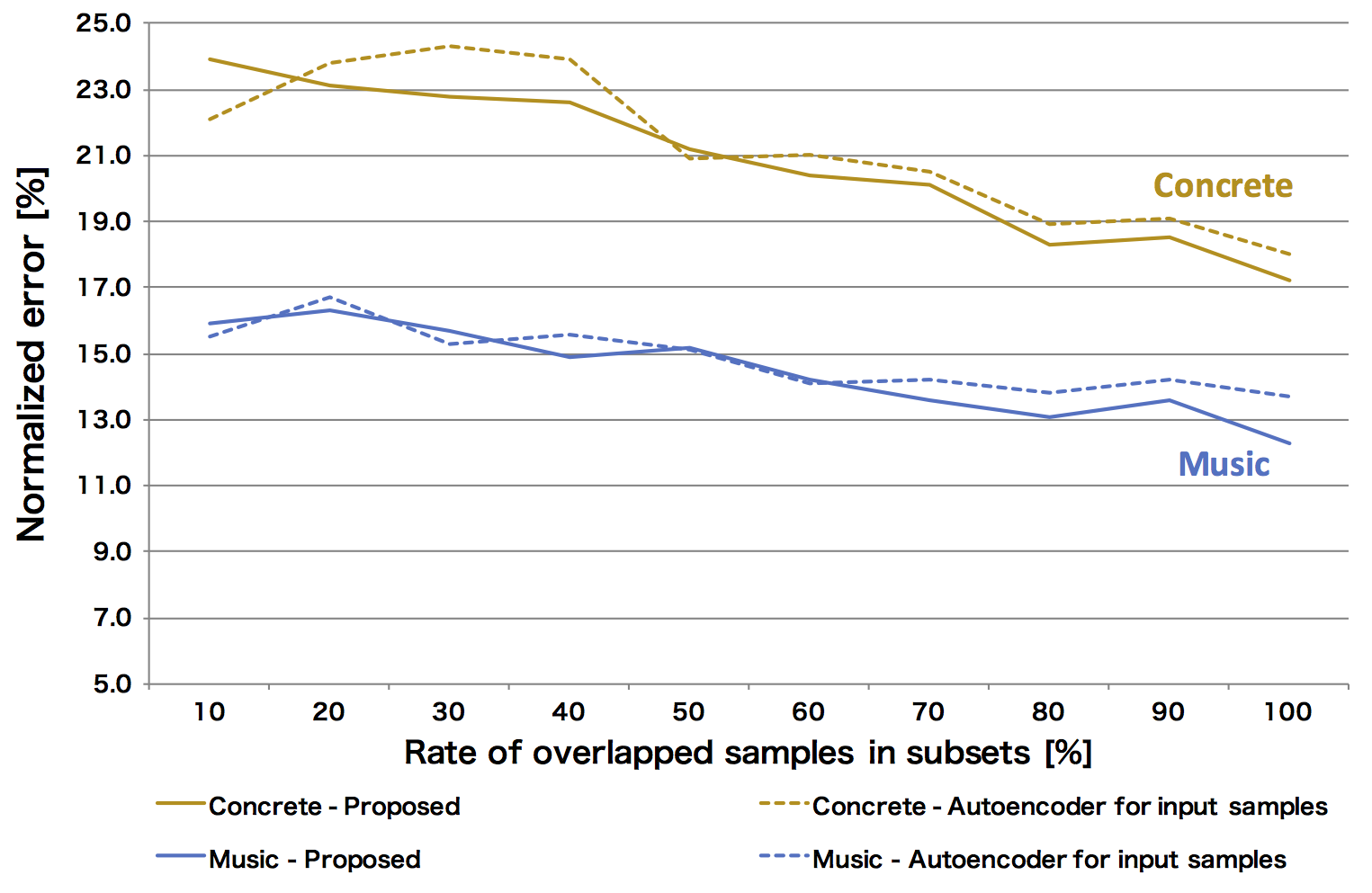}
    \includegraphics[mediaboxonly,width=0.485\columnwidth]{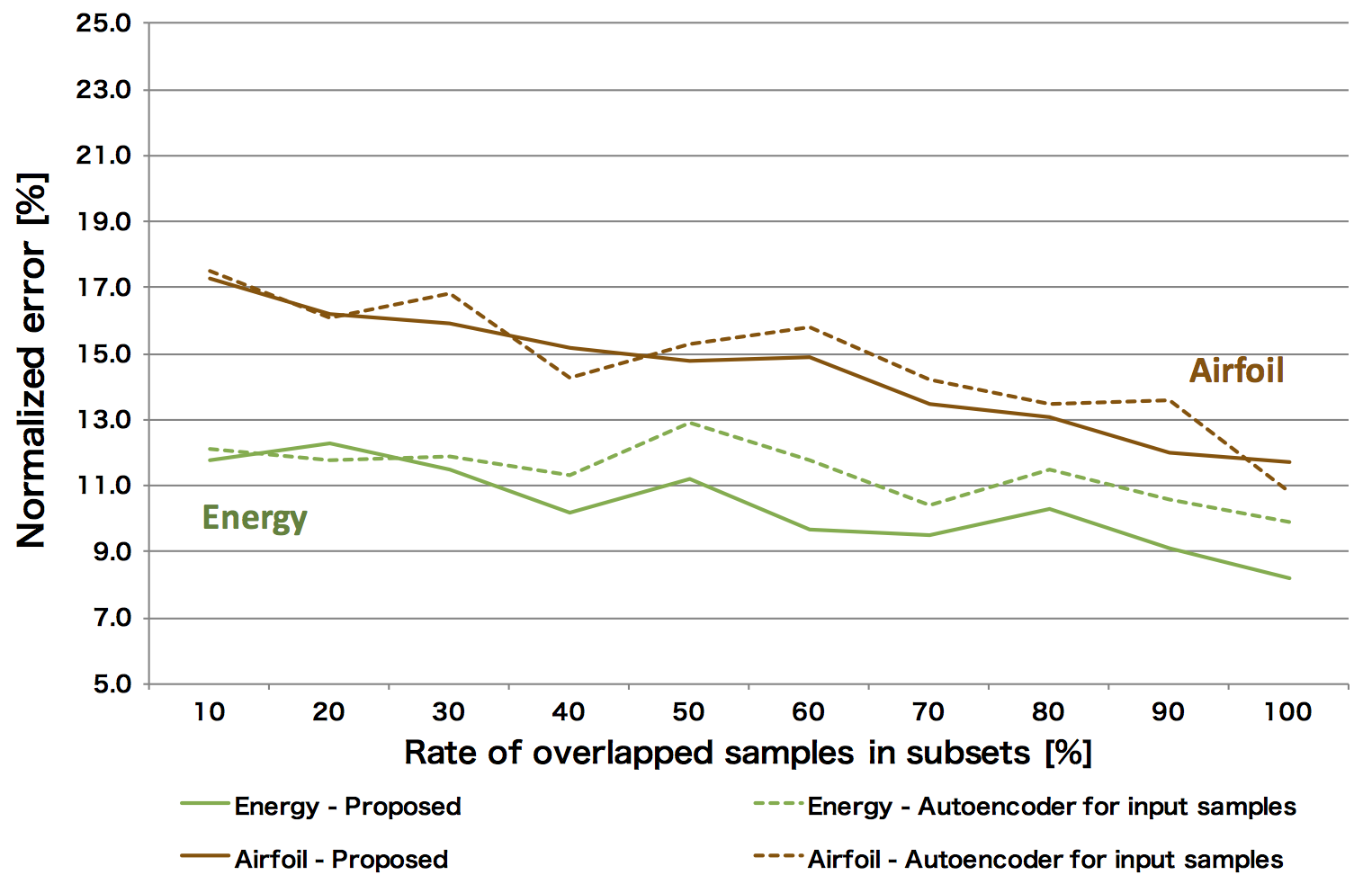}
    \caption{%
      Effectiveness of the proposed method applying denoising autoencoders
      to indicating vectors, for (left) Concrete and Music datasets, and
      (right) Airfoil and Energy datasets.
    }
    \label{fig:applyEncoder}
  \end{center}
\end{figure}

You may think that denoising autoencoders should be applied to input features, not traversal patterns.
We here answer this concern by comparing the regression performances of denoising autoencoders applying to test features (baseline) and indicator vectors (proposed).
Figure \ref{fig:applyEncoder} show the experimental results.
We confirmed that our proposed method that applies denoising autoencoders to correcting traversal paths outperformed the method that applied denoising autoencoders to removing noises from input features for many datasets, although the gain was not so large ($1.6$\% on average over all the datasets).
In particular, we had about 2.5 points improvement for the Energy dataset when the subset overlap ratio is 60\%.

\subsection{Effect of tree selection by cross entropy}
\label{sec:exp:select}

\begin{table}[t]
  \scriptsize
  \centering
  \caption{Effect of regression-tree selection by cross entropy}
  \label{tab:hh8}
  \begin{tabular}{c|c|c|c|c|c|c} \hline
    & \multicolumn{2}{c|}{Cross entropy}&\multicolumn{2}{c}{Noise detection rate[\%]}& \multicolumn{2}{|c}{Regression error[\%]}  \\ \hline
    & w/ selection & w/o selection & w/ selection & w/o selection & w/ selection & w/o selection \\ \hline
    Concrete & 0.64 & 0.82 & 71.2 & 68.1 & 20.4 & 22..4 \\ \hline
    Music    & 0.62 & 0.74 & 72.9 & 71.4 & 14.2 & 15.7 \\ \hline
    Energy   & 0.68 & 0.91 & 67.2 & 63.1 & 9.7 & 12.9 \\ \hline
    Airfoil  & 0.55 & 0.66 & 68.9 & 67.7 & 14.9 & 16.3 \\ \hline\hline
    Average&{\bf 0.62} & 0.78& {\bf 70.0}&67.6&{\bf14.8}&16.8 \\ \hline
  \end{tabular}
\end{table}

Finally, we evaluated the effectiveness of selecting trees on the basis of cross entropy.
As measures for evaluation, we used the precision rate and the cross entropy and used in Section \ref{sec:exp:denoising} for evaluating noise correction performances and the normalized $L_1$-distance used in Section \ref{sec:exp:regression} for evaluating regression performances.
Table \ref{tab:hh8} shows the result.
This indicates that the proposed tree selection approach made it easier to find and correct noise patterns with denoising autoencoders.
Some may suppose that the fact that proposed tree selection approach leaves only mutually correlated trees may causes over-fitting. However, the experimental results shows that the proposed method effectively improved the regression performance.

\section{Conclusion}
\label{sec:conclude}

In this paper, we proposed a novel method for boosting the resistance of random forests to noises in test samples by effectively incorporating denoising autoencoders.
We employed indicator vectors each of which indicates leaf nodes where an input sample has arrived.
Our proposed method called denoising random forests employed denoising autoencoders to detect and correct noise-corrupted decision nodes from indicator vectors.
Estimation considering multiples traversal paths eased the adversarial effects caused by incorrect estimation of denoising autoencoders.
Various types of experimental results demonstrated the effectiveness of the proposed method.
Promising future work includes (1) the improvement of recovering true traversal paths by introducing emerging deep learning technologies and (2) the extension of the proposed method to other machine learning tasks such classification and density estimation.




\bibliographystyle{elsarticle-harv}
\bibliography{hibino}




\end{document}